%% file: main.tex
\documentclass[10pt,twocolumn,letterpaper]{article}

\usepackage[pagenumbers]{cvpr} 

\input{preamble}

\definecolor{cvprblue}{rgb}{0.21,0.49,0.74}
\usepackage[pagebackref,breaklinks,colorlinks,citecolor=cvprblue]{hyperref}
\usepackage[accsupp]{axessibility}
\usepackage{colortbl}
\usepackage{rotating}
\usepackage{multirow}
\usepackage{graphicx}
\usepackage{pifont}
\usepackage{tcolorbox}
\tcbuselibrary{most}

\definecolor{lightgray}{gray}{0.9}
\definecolor{lightorange}{RGB}{255, 204, 153} 

\definecolor{lightgreen}{RGB}{32, 144, 140}  
\definecolor{yellow}{RGB}{68, 190, 112} 
\definecolor{yellowgreen}{RGB}{189, 222, 38} 
\definecolor{darkblue}{RGB}{52,94,141} 

\definecolor{c1}{RGB}{180, 242, 230}  
\definecolor{c2}{RGB}{137, 203, 225} 
\definecolor{c3}{RGB}{127, 168, 216} 
\definecolor{c4}{RGB}{120,135,204} 
\def\paperID{13512} 
\def\confName{CVPR}
\def\confYear{2025}

\newcommand{\finding}[2]{
    \vspace{-0.1cm}
    \begin{tcolorbox}[
        colback=white!98!brown,     
        colframe=teal!60!black,     
        arc=5pt,                    
        boxsep=5pt,                 
        left=10pt,                  
        right=10pt,                 
        top=2pt,                    
        bottom=2pt,                 
        boxrule=0.8pt,              
        drop shadow=gray!50!white,  
        enhanced jigsaw             
    ]
    \vspace{-0.1cm}
        \paragraph{\textbf{\textit{Finding #1:}}} #2
    \vspace{-0.1cm}
    \end{tcolorbox}
    \vspace{-0.3cm}
}

\title{Assessing and Learning Alignment of Unimodal Vision and Language Models}

\author{
  \textbf{Le Zhang}\textsuperscript{1,2} \hspace{1.5em} 
  \textbf{Qian Yang}\textsuperscript{1,2} \hspace{1.5em} 
  \textbf{Aishwarya Agrawal}\textsuperscript{1,2,3} \\
  \small \textsuperscript{1}Mila - Quebec AI Institute, \textsuperscript{2}Université de Montréal, \textsuperscript{3}Canada CIFAR AI Chair
}

\begin{document}
\maketitle
\input{sec/abstract}

\input{sec/intro}

\input{sec/linearasses}

\input{sec/sail}

\input{sec/relatedwork}

\input{sec/conclusion}

    \section*{Acknowledgement}
We sincerely appreciate the valuable feedback provided by Rabiul Awal, Saba Ahmadi, and Oscar Mañas, as well as the thoughtful input from all MAIR Lab members on multiple occasions. We thank the Mila IDT team and their technical support for maintaining the Mila compute cluster. We also acknowledge the material support of NVIDIA in the form of computational resources. Throughout this project, Aishwarya Agrawal received support from the Canada CIFAR AI Chair award.

{
    \small
    \bibliographystyle{ieeenat_fullname}
    \bibliography{main}
}

\begin{appendix}
\input{sec/X_suppl}
\end{appendix}

\end{document}

%% file: preamble.tex
\usepackage[dvipsnames]{xcolor}


%% file: sec/abstract.tex
\begin{abstract}
How well are unimodal vision and language models aligned? 
Although prior work have approached answering this question, their assessment methods do not directly translate to how these models are used in practical vision-language tasks. 
In this paper, we propose a direct assessment method, inspired by linear probing, to assess vision-language alignment. 
We identify that the degree of alignment of the SSL vision models depends on their SSL training objective, and we find that the clustering quality of SSL representations has a stronger impact on alignment performance than their linear separability. Next, we introduce Swift Alignment of Image and Language (SAIL), a efficient transfer learning framework that aligns pretrained unimodal vision and language models for downstream vision-language tasks. Since SAIL leverages the strengths of pretrained unimodal models, it requires significantly fewer ($\sim$6\%) paired image-text data for the multimodal alignment compared to models like CLIP which are trained from scratch. SAIL training only requires a single A100 GPU, $\sim$5 hours of training and can accommodate a batch size up to 32,768. SAIL achieves 73.4\% zero-shot accuracy on ImageNet (vs. CLIP's 72.7\%) and excels in zero-shot retrieval, complex reasoning, and semantic segmentation. Additionally, SAIL improves the language-compatibility of vision encoders that in turn enhance the performance of multimodal large language models.\footnote{Project page: \href{https://lezhang7.github.io/sail.github.io/}{https://lezhang7.github.io/sail.github.io/}}


\end{abstract}

%% file: sec/intro.tex
\section{Introduction}
\label{sec:intro}

\begin{figure}[!ht]
    \centering
    \includegraphics[width=\linewidth]{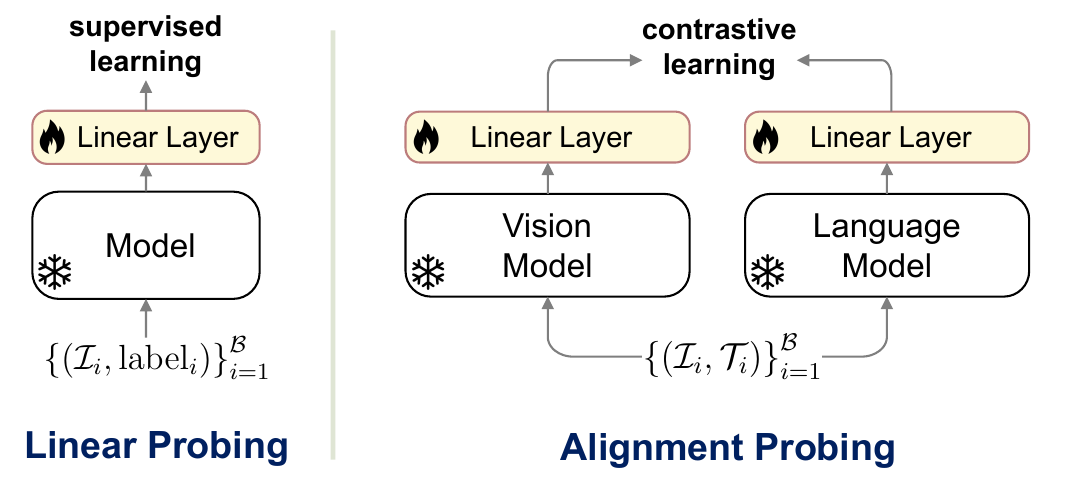}
\caption{\textbf{Conceptual Overview}: Alignment probing evaluates the alignment potential of two pretrained uni-modal models. }
    \label{fig:conceptualdiff}
    \vspace{-5mm}
\end{figure}

The integration of language and vision is pivotal in advancing models for computer vision tasks \cite{llava, liu2024improved, sam, lai2024lisa, zhang2023moqagpt}. This raises a key question: \textit{“To what extent can unimodal visual and language models be aligned with each other?"} In particular, how do unimodal representations impact cross-modal alignment: Do larger unimodal models trained on extensive datasets yield better cross-modal alignment? Does the choice of self-supervised learning (SSL) method play a critical role in determining the alignment strength? What property of SSL representation correlates the most with cross-modal alignment performance: is it linear separability or the clustering quality?


Although prior work have approached answering the question of \textit{``To what extent are pretrained unimodal models aligned with each other"}, their assessment methods do not directly translate to how these models are used in practical vision-language tasks. For instance, \citet{huh2024platonic} assess cross-modal alignment using mutual nearest-neighbor metrics, indicating a certain level of alignment across models trained on separate modalities. However, this assessment method serves only as a proxy method as it focuses on relative ordering within each modality rather than directly measuring cross-modal distances. The latter is precisely how inference with vision-language models (VLMs) is conducted, so we believe directly measuring cross-modal distances is a more direct measurement of alignment performance.  

To quantitatively answer aforementioned questions regarding the impact of modality-specific representation on cross-modal alignment, we introduce \textbf{visual-language alignment probing}, akin to \textbf{linear probing} used in SSL evaluation. As illustrated in \cref{fig:conceptualdiff}, alignment probing freezes the pretrained vision and language backbones and trains a lightweight \emph{linear} alignment layer on image-text datasets. We evaluate alignment through zero-shot retrieval tasks, finding that vision and language models exhibit strong alignment generally. However, the degree of alignment of the SSL vision models depends on their SSL training objective. We also found that the clustering quality of visual representations, as indicated by k-NN classifier performance, has a stronger impact on image-text alignment performance than their linear separability. Furthermore, for complex visio-linguistic reasoning, strong language understanding is essential. CLIP training even with scaled model and training data size is insufficient for developing a high-quality text encoder.

Building on the findings that unimodal vision and language models show inherent alignment, and that language models trained on extensive natural language data serve as effective text encoders for VLMs, we propose Swift Alignment of Image and Language (SAIL) for learning better vision-language alignment. SAIL is an efficient transfer learning framework designed to construct robust foundational VLMs leveraging high-quality pretrained unimodal vision and language models. To enhance alignment quality, we employ three optimized components: a non-linear alignment layer, a refined contrastive loss function, and MLLM-generated high-quality training captions. SAIL is highly data-efficient, leveraging pretrained unimodal models and requiring only about 6\% of the paired image-text data needed for models like CLIP, which are trained from scratch. It’s also compute-efficient, needing only a single A100 GPU, $\sim$5 hours of training, and supporting batch sizes up to 32,768 by training just the alignment layer.


 By aligning the pretrained vision-encoder DINOv2-L and the pretrained language encoder NV2 using 23M image-text pairs, our method SAIL outperforms CLIP trained on 400M image-text pairs by 0.7\% on ImageNet, and by 5.6\% and 2.7\% on COCO text-to-image and image-to-text retrieval respectively. SAIL leverages the strengths of unimodal models -- DINOv2’s fine-grained visual understanding and NV2’s complex language reasoning -- and  excels considerably  in challenging vision-language tasks such as Winoground \cite{thrush2022winoground} and MMVP \cite{tong2024eyes}, as well as in open-vocabulary image segmentation tasks. Furthermore, in comparison with prior efficient training methods like LiT \cite{zhai2022lit} and Sharelock \cite{sharelock}, which focus on tuning language models to align with frozen vision encoders, SAIL improves both the alignment performance as well as the language-compatibility of the vision encoder itself. This enables SAIL’s vision encoder to be transferable to MLLMs, resulting in 
 significant performance gains; when integrated with LLaVA-1.5 \cite{llava}, SAIL’s alignment training pushes the capabilities of the DINOv2 vision encoder from lagging behind the CLIP vision encoder to surpassing it in 5 out of 7 tasks downstream MLLM tasks.

%% file: sec/linearasses.tex
\begin{figure*}[!th]
    \centering
    \begin{minipage}[b]{\linewidth}  
        \centering
        \begin{minipage}[b]{0.48\linewidth}
            \centering
            \includegraphics[width=\linewidth]{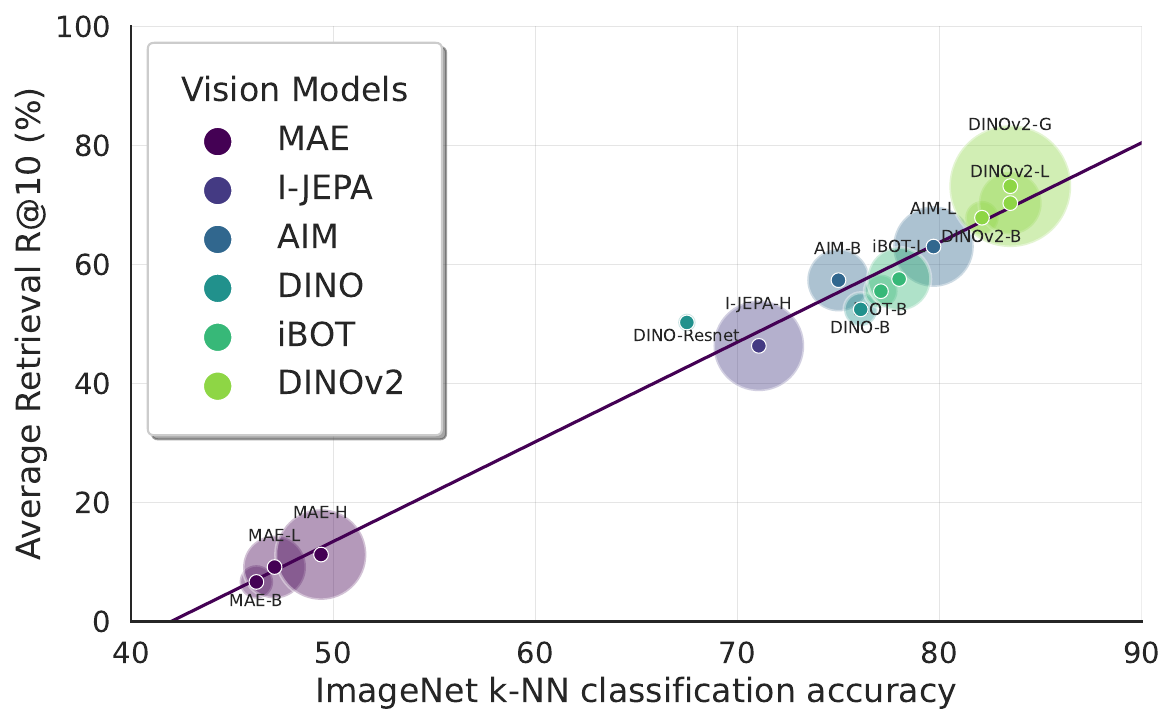}
            \label{fig:visionalign1}
        \end{minipage}
        \hspace{0.01\linewidth}  
        \begin{minipage}[b]{0.48\linewidth}
            \centering
            \includegraphics[width=\linewidth]{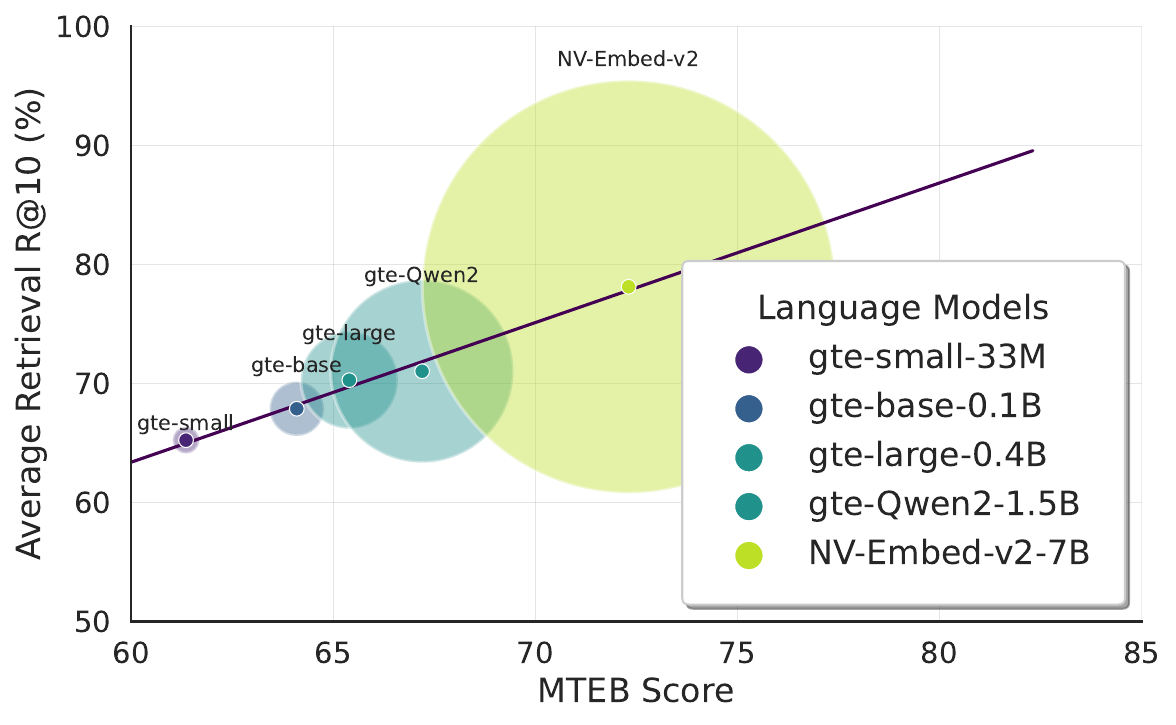}
            \label{fig:visionalign2}
        \end{minipage}
    \end{minipage}
    \caption{\textbf{Linear alignment probing results trained with 2.2M paired data from CC3M.} The radius represents the \textbf{relative number of parameters} in each model. The Y-axis indicates the zero-shot MSCOCO retrieval average R@10 performance. (Left) the X-axis shows kNN performance for various SSL models. (Right) the X-axis displays MTEB average scores across models.}
    \label{fig:alignmentresults}
      \vspace{-3mm}
    
\end{figure*}

\section{Part I: Assessing Alignment between Unimodal Models}

\subsection{Approach and Experimental Setup}


In this section, we evaluate the alignment potential of pretrained unimodal models to determine those most compatible with models of the other modality. To focus on the alignment capacity of the pretrained models, we use a \emph{linear} alignment layer to connect their representations. We refer to this as \emph{alignment probing}. We use linear layers instead of MLPs because the latter could introduce additional alignment capability and hence confound the findings. The alignment probing architecture is illustrated in \cref{fig:conceptualdiff}: only the alignment layer is trained, while the backbones remain frozen. We employ contrastive learning to pull matched image-text pairs closer and push unmatched pairs further in the representation space; see Appendix for training details.

We use the open-source CC3M dataset (2.2M paired image-text samples) to train the alignment layer, leveraging its diversity and quality as an effective probing dataset. To measure the alignment quality, we test on COCO in zero-shot retrieval setup, using the R@10 metric. We report average recall of text-to-image and image-to-text retrieval tasks.

For systematic evaluation, we fix an anchor model in one modality and vary models in the other modality to identify which models best align with the anchor. For the language anchor, we select \textit{GTE-en-large-v1.5} due to its robust performance across language understanding tasks \cite{muennighoff2022mteb}. We evaluate a range of vision models with this anchor, including various SSL methodologies: masked image modeling with discrete tokenizers (e.g., iBOT \cite{zhou2021ibot}), pixel-level reconstruction (e.g., MAE \cite{mae}), knowledge distillation (e.g., DINO \cite{dino} and DINOv2 \cite{oquab2023dinov2}), and autoregressive image modeling (e.g., AIM \cite{aim}). Additionally, we incorporate a ResNet \cite{resnet} model trained with DINO to assess the effect of architectural variations.

Similarly, we use \textit{DINOv2-Large} as the vision anchor and evaluate a range of language models, including encoder-only models like GTE-en-large-v1.5 \cite{gte} and decoder-only models, such as the GTE-Qwen2 \cite{zhang2024mgte} and NV-Embed-v2 \cite{lee2024nv}. This combination allows us to systematically analyze how pretrained  vision models and language models contribute to cross-modal alignment across different architectures and pretraining strategies.

\subsection{Results and Findings}

\paragraph{Language as Anchor.} 

The main results are displayed in the left panel of \cref{fig:alignmentresults}. Using our proposed alignment metric, we observe that most models achieve strong performance, with Retrieval R@10 ranging from 50\% to 75\%. This suggests a global alignment exists between unimodal vision and language models, consistent with observations from previous studies \cite{huh2024platonic}. Notably, DINOv2 outperform all other SSL models, achieving the strongest alignment with the language anchor. Surprisingly, AIM-L, despite its 1 billion parameters, underperforms  DINOv2-B, which has only a 86M parameters. Within the same training framework, DINO-ResNet achieves performance comparable to DINO-B with fewer parameters, indicating the high effectiveness of ResNet in alignment tasks.

MAE-series models, on the other hand, exhibit markedly weaker alignment compared to other models. This may stem from their pixel-level reconstruction SSL objective, which focuses on low-level details (reconstructing each pixel perfectly) rather than the high-level semantics essential for the image-text alignment tasks. Additionally, model size positively impacts alignment performance, with larger models consistently yielding better alignment outcomes.

We further investigate which properties of SSL representations best support alignment. There are two standard metrics to probe the quality of SSL representations \cite{moco, simclr, dino, oquab2023dinov2}: 1) k-NN classifier, which measures  \emph{non-linear} separability and clustering quality of the SSL representations, i.e., how well representations of the same concept \emph{cluster} together, 2) Linear Probing which measures \emph{linear separability}, i.e., how effectively the features can be separated by a linear boundary. Our analysis reveals a strong linear correlation between k-NN performance and alignment score, as illustrated by an approximated line in the figure. In contrast, Linear Probing (see \cref{fig:visionalign_linearprobe} in Supp.) shows a weaker correlation with alignment scores . Specifically, Pearson’s correlation between alignment accuracy (computed using our proposed metric) and ImageNet classification accuracy is 0.991 for k-NN classification and 0.847 for linear probing, highlighting that non-linear separability (i.e., clustering quality) matters more than linear-separability for image-text alignment.

\finding{1}{Alignment performance strongly depends on the clustering quality of SSL representation, as reflected by k-NN performance.}

 \paragraph{Vision as Anchor.} Our results in \cref{fig:alignmentresults} (right panel) show that the MTEB \cite{muennighoff2022mteb} average score, measuring performance across 56 language understanding tasks, correlates almost linearly with alignment scores, with a Pearson correlation of 0.994. This suggests that the language understanding capability is critical for image-text alignment performance.
 
We observe a clear trend: stronger language models (measured by MTEB) consistently yield better alignment with the vision anchor. Notably, the compact gte-small achieves 80\% of CLIP’s alignment performance with only 30\% of CLIP's text encoder parameter count; and NV-Embed-2 \cite{oquab2023dinov2} reaches alignment scores comparable to CLIP-L (78.1\% vs. 80.1\%) while training on a much smaller dataset (2.2M vs. 400M pairs). This underscores the strength of models trained on natural language data in aligning text semantics into a shared representation space.
 
\begin{figure}[tb]
    \centering
    \includegraphics[width=\linewidth]{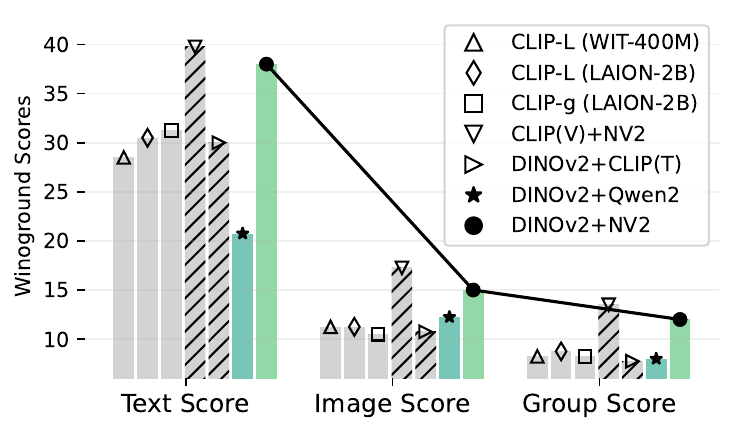}
    \captionsetup{font=small}
    \caption{\textbf{Winoground Results.} CLIP(V) represents vision encoder and CLIP(T) represents text encoder from CLIP-L(WIT-400M). `+' indicates alignment probing with two models.}
    \label{fig:complex reasoning}
    \vspace{-3mm}
\end{figure}

        

Additionally, we identify that the language understanding significantly influences vision-language complex reasoning. \cref{fig:complex reasoning} demonstrates this in Winoground task \cite{thrush2022winoground}: although CLIP is scaled in both data (400M $\rightarrow$ 2B) and model size (427M $\rightarrow$ 1366M), the scores remains low in metrics like “Image” (11.25) and “Group” (8.25) scores\footnote{See appendix for more details about the metric.}. This outcome suggests that merely scaling up CLIP training dataset and model size may not be sufficient for enhancing complex reasoning. A potential explanation could be that CLIP’s training data, primarily web-sourced descriptions, may lack the rich semantic information necessary to learn advanced reasoning for its text encoder. 

In contrast, a stronger language model like NV-Embed-v2, extensively trained on a vast corpus of text, significantly boosts performance in complex vision-language reasoning, despite using limited alignment data (2.2M): replacing the CLIP text encoder with NV2 yields substantial improvements over the original CLIP; similarly, when paired with DINOv2-L vision encoder, NV2 significantly outperforms both CLIP-L text encoder and the Qwen2-1.5B model (\cref{fig:complex reasoning})  suggesting that:

 \finding{2}{Language understanding is key for complex vision-language reasoning. CLIP training alone is insufficient to learn a good enough text encoder. Leveraging rich pool of pretrained language models as text encoders offers a promising approach to building robust foundation VLMs.}





%% file: sec/sail.tex
\section{Part II: Learning Alignment between Unimodal Models}

\subsection{Swift Alignment of Image and Language} The assessment results demonstrate that unimodal vision and language models are inherently aligned, and reveals that foundational VLMs such as CLIP can significantly benefit from using strong pretrained text encoders. To build powerful CLIP-like models harnessing the strengths of robust off-the-shelf unimodal vision and language models, we introduce Swift Alignment of Image and Language (SAIL): an efficient transfer learning framework that transfers learned unimodal visual and textual representation to downstream vision-language tasks. SAIL improves alignment through three optimized components: (1) alignment layer architecture, (2) loss function, and (3) data quality. We conduct ablation studies to validate the effectiveness of the choices. 

For all experiments, we use DINOv2-L as the  vision model and GTE-en-large-v1.5 as the language model based on their compact model size and good alignment potential. We train with CC3M as the base set-up and evaluate on ImageNet-1k and COCO. Ablation results are shown in \cref{tab:sail_abl}.

\paragraph{Alignment Layer.} The alignment layer $\mathcal{G}(\cdot)$ plays a crucial role in aligning modality-specific features in frozen vision and language encoders with each other. Our experiments demonstrate that using a single-layer, non-linear Gated Linear Unit (GLU) \cite{shazeer2020glu} with ReLU activation significantly improves alignment compared to baselines that use linear layers. As shown in \cref{tab:sail_abl}, replacing the linear layer (row 0) used in LiT \cite{zhai2022lit} with the MLP from ShareLock \cite{sharelock} improves classification but reduces retrieval performance. In contrast, GLU layers consistently improve performance on all tasks, with a GLU $\times 8$ boosting top-1 accuracy on IN-1K by 12.2\%, T2I R@1 by 5\%, and I2T R@1 by 9\%. By limiting tunable parameters to this lightweight GLU layer with minimal FLOPs, we achieve efficient and targeted optimization across vision-language tasks.

\begin{table}[!tb]
     \centering
    \resizebox{\linewidth}{!}{%
      \begin{tabular}{lccccc}

\textbf{Method} &  \textbf{IN-1K 0-shot} & \textbf{T2I R@1} & \textbf{I2T R@1} \\
\toprule
\rowcolor{lightgray!60}
0\ \ \  Baseline  &  33.2 & 11.1 & 13.5\\
\rowcolor{lightorange!40}   
1\ \ \  + MLP $\times 4$   &  36.8 & 8.0  & 10.7 \\
\rowcolor{lightorange!40}   
2\ \ \  + GLU $\times 4$     &  39.6 & 11.5 & 17.4 \\  
\rowcolor{lightorange!40}
3\ \ \  + GLU $\times 8$   & 45.4  & 16.1 & 22.5 \\
\rowcolor{yellow!20}
4\ \ \  + Sigmoid & 50.7 & 25.4 & 36.0\\
\rowcolor{yellow!20}
5\ \ \  \quad +$|\mathcal{B}| \rightarrow |\mathcal{B}|^2$ & 51.8 & 26.2 & 36.7 \\

\rowcolor{lightgreen!20}
6\ \ \  + Long-HQ  &  48.4 & 31.4 & 44.2\\
\rowcolor{lightgreen!20}
7\ \ \  + Multi-Pos &  54.0 & 32.9 & 45.4 \\

\end{tabular}
    }
    \captionsetup{font=small}

  \caption{\textbf{Ablation results using CC3M} on \colorbox{lightorange!40} {Alignment Layers}, \colorbox{yellow!20}{Loss} and \colorbox{lightgreen!20}{Data}. Baseline refers to aligning unimodal models with only linear layer using infoNCE loss \cite{clip}. +'s indicate addition of the component on top of the immediately previous row. $\times n$ represents an intermediate dimensionality scaled by $n$ times the input dimensionality. IN-1K refer to zero-shot top1 accuracy on ImageNet-1k; text-to-image(T2I) and image-to-text(I2T) refer to retrieval results on COCO.}
    \label{tab:sail_abl}
      \vspace{-3mm}
\end{table}

\paragraph{Contrastive Loss.} SAIL samples a batch of image-text pairs $\{ (\mathcal{I}_i, \mathcal{T}_i) \}_{i=1}^\mathcal{B}$, processed through image encoders $\mathcal{F}_I(\cdot)$ and text encoder $\mathcal{F}_T(\cdot)$ and their alignment layers $\mathcal{G}_I(\cdot)$ and $\mathcal{G}_T(\cdot)$. To improve compute-efficiency and performance, we use the binary classification-based Sigmoid loss \cite{siglip} instead of CLIP's InfoNCE. This approach reduces the computational overhead of softmax normalization and enhances the model's sensitivity to hard negatives. The loss is defined as:

\begin{equation}
    \begin{aligned}
        \mathcal{L}(\mathcal{I}, \mathcal{T}) = -\frac{1}{|\mathcal{B}|} \sum_{i=1}^{|\mathcal{B}|} \sum_{j=1}^{|\mathcal{B}|} \log \frac{1}{1 + e^{z_{ij}(-t \hat{\mathbf{x}}_i \cdot \hat{\mathbf{y}}_j + b)}},
    \end{aligned}
\end{equation}

where \( \mathbf{x}_i = \mathcal{G}_I(\mathcal{F}_I(\mathcal{I}_i)) \) and \( \mathbf{y}_i = \mathcal{G}_T(\mathcal{F}_T(\mathcal{T}_i)) \). Each feature is then L2-normalized as \( \hat{\mathbf{x}}_i = \frac{\mathbf{x}_i}{\|\mathbf{x}_i\|_2} \) and \( \hat{\mathbf{y}}_j = \frac{\mathbf{y}_j}{\|\mathbf{y}_j\|_2} \).  The similarity score, \( s_{ij} = -t \hat{\mathbf{x}}_i \cdot \hat{\mathbf{y}}_j + b \), incorporates temperature scaling ($t$) and bias ($b$), with $z_{ij}$ set to 1 if $i = j$ and -1 otherwise. As shown in \cref{tab:sail_abl}, using sigmoid loss (row 4) significantly outperforms InfoNCE (row 3) across all tasks, with gains of 5.3\% on ImageNet-1k, 9.3\% on T2I, and 13.5\% on I2T for COCO.

Furthermore, we find that averaging the loss across all pairs—rather than just the positive pairs (i.e., replacing $|\mathcal{B}|$ in row 4 with $|\mathcal{B}|^2$ in row 5)—ensures equal contribution from both positive and negative samples, which leads to performance improvements, with gains of 1.1\% on ImageNet-1k, 1.5\% on T2I, and 1.2\% on I2T for COCO.

\begin{table*}[!ht]
\centering
\resizebox{\textwidth}{!}{%
\begin{tabular}{ll|rrrrrrrrrrr|r}
\toprule
\textbf{Data} & \textbf{Model} & Food101 & CIFAR10 & CIFAR100 & SUN397 & Cars & Aircraft & DTD & Pets & Cal101 & Flowers & Avg. & \textbf{IN-1K} \\

\midrule
\multicolumn{14}{c}{\textit{Model Architecture: ViT-B/16}}\\ \midrule
\multirow{4}{*}{CC3M} 
& CLIP-B\cite{zheng2024dreamlip} & 10.6 & 53.9 & 20.4 & 31.2 & 1.2 & 1.1 & 10.4 & 11.7 & 43.2 & 12.9 & 19.7 & 16.0  \\
& DreamLIP\cite{zheng2024dreamlip} & 19.4 &74.3& 44.2& 45.9& 2.8& 1.0& 17.0& 27.1& 63.1& 14.7& 31.0& 31.1 \\
& LiT\cite{sharelock} & -&-& -& - & 3.0& 2.1 & - & 28.5 & - & 35.9 & - & 44.1\\
& SAIL-B-GTE & 47.1&94.1&74.6&63.9&9.2&4.2&49.7&39.5&77.9&31.8&49.2 & 50.7\\ 
\midrule
\multirow{6}{*}{CC12M} & CLIP-B\cite{zheng2024dreamlip} & 25.3&66.5&32.1&39.9&14.7&1.9&13.5&45.0&59.8&15.0&31.4&34.0 \\
& DreamLIP\cite{zheng2024dreamlip} & 58.3&87.3&62.6&54.3&29.7&4.9&29.2&60.3&83.1&28.9&49.9&50.3 \\
& LiT\cite{sharelock} & -&-& -& - & 13.2& 5.0 & - & 74.4 & - & 48.2 & - & 56.2 \\
& ShareLock$\dag$\cite{sharelock} & -&-& -& - & 11.5& 8.3 & - & 66.6 & - & 48.8 & - & 59.1 \\
& SAIL-B-GTE$\dag$ & 63.1&\cellcolor{lightorange!40}94.1&\cellcolor{lightorange!15}78.2&64.2&28.1&6.6&52.0&60.1&81.5&49.4&57.7&58.7\\ 
& SAIL-B-NV2$\dag$ & 77.7&\cellcolor{lightorange!15}93.8&\cellcolor{lightorange!40}79.9&66.2&35.8&13.4&\cellcolor{lightorange!40}61.5&81.7&82.1&61.5&65.4&\cellcolor{lightorange!40}68.1\\ 
\rowcolor{gray!15}
\textcolor{gray!70}{LAION400M} & \textcolor{gray!70}{CLIP-B} & \textcolor{gray!70}{85.5} & \textcolor{gray!70}{93.0} & \textcolor{gray!70}{71.7} & \textcolor{gray!70}{66.8} & \textcolor{gray!70}{83.5} & \textcolor{gray!70}{16.7} & \textcolor{gray!70}{52.8} & \textcolor{gray!70}{90.1} & \textcolor{gray!70}{91.2} & \textcolor{gray!70}{63.9} & \textcolor{gray!70}{65.5} & \textcolor{gray!70}{67.0} \\ \midrule

\multicolumn{14}{c}{\textit{Model Architecture: ViT-L/14 }}\\ \midrule
CC12M & SAIL-L-GTE & 71.2&\cellcolor{lightorange!15}96.3&\cellcolor{lightorange!15}83.8&67.2&33.0&8.0&53.0&66.5&\cellcolor{lightorange!15}82.6&57.7&61.9&63.9\\
23M Merged & SAIL-L-GTE & 76.1&\cellcolor{lightorange!15}97.3&\cellcolor{lightorange!15}84.6&68.6&32.0&16.0&52.5&56.9&\cellcolor{lightorange!15}83.0&68.3&63.5&65.4 \\

CC12M & SAIL-L-NV2 & 81.9&\cellcolor{lightorange!15}96.1&\cellcolor{lightorange!15}85.2&68.3&42.9&16.3&\cellcolor{lightorange!15}60.4&84.7&\cellcolor{lightorange!15}82.4&67.5&68.6&72.1 \\
23M Merged & SAIL-L-NV2 & 86.1&\cellcolor{lightorange!40}96.7&\cellcolor{lightorange!40}86.7&69.8&44.6&\cellcolor{lightorange!40}28.6&\cellcolor{lightorange!40}63.5&82.3&\cellcolor{lightorange!40}85.4&\cellcolor{lightorange!40}77.2&72.1&\cellcolor{lightorange!40}73.4\\

\rowcolor{gray!15}
\textcolor{gray!70}{LAION400M} & \textcolor{gray!70}{CLIP-L} & \textcolor{gray!70}{90.1}&\textcolor{gray!70}{94.6}&\textcolor{gray!70}{77.4}&\textcolor{gray!70}{72.6}&\textcolor{gray!70}{89.6}&\textcolor{gray!70}{25}&\textcolor{gray!70}{60.4}&\textcolor{gray!70}{91.7}&\textcolor{gray!70}{82.1}&\textcolor{gray!70}{75.5}&\textcolor{gray!70}{75.9}&\textcolor{gray!70}{72.7}
 \\

\bottomrule
\end{tabular}}

\captionsetup{font=small}
    \caption{\textbf{Zero-shot Image Classification top1 accruacy}. CC3M contains 2.2 million samples, while CC12M includes 7.7 million samples. $\dag$Note that the patch size for DINOv2-B is 14. We highlight the models with \colorbox{lightorange!40}{best performance} and \colorbox{lightorange!25}{better than CLIP} among the models using the same vision encoder architecture. \colorbox{lightgray!60}{CLIP trained on larger LAION400M dataset is provided as reference.} }
\label{tab:image classification}
\vspace{-3mm}
\end{table*}

\paragraph{High-Quality Data.} Recent work indicates that VLMs benefit from training on smaller, higher-quality datasets \cite{li2023blip, llava, zheng2024dreamlip, lalip}.  As shown in \cref{tab:sail_abl}, raw web-collected short captions that focus on a single object (row 5) are beneficial for image classification. In contrast, longer, high-quality synthetic captions—such as those generated with ShareGPT4 \cite{chen2023sharegpt4v} for each image in CC3M (row 6)—boost performance on retrieval tasks requiring nuanced visio-linguistic understanding, though they are less effective for object recognition.

To leverage both benefits, we combine long and short captions within each training batch, offering diverse training signals that enhance representation learning and task adaptability (row 7). For each image-caption pair $\{ (\mathcal{I}_i, \mathcal{T}_i) \}$, we include high-quality synthetic caption $\mathcal{T}_i^{HQ}$ as additional positives. The multiple positive caption contrast is then defined as:

\begin{equation}
    \begin{aligned}
        \mathcal{L}_\text{Multi-Pos} = \mathcal{L}(\mathcal{I}, \mathcal{T}) + \mathcal{L}(\mathcal{I}, \mathcal{T^{\text{HQ}}}),
    \end{aligned}
\end{equation}

\begin{figure}[!tbh]
    \centering
    \includegraphics[width=\linewidth]{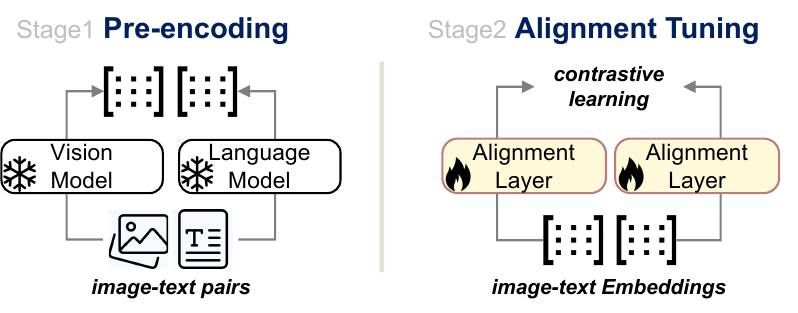}
\caption{\textbf{SAIL Pipeline.} Image-Text data is pre-encoded into embeddings. During alignment tuning, only embeddings and alignment layers are loaded to reduce GPU memory consumption and accelerate training speed.}
    \label{fig:trainingpipe}
\end{figure}

\noindent\textbf{Cheap Training Recipe.}\label{sec:cheaptrain} SAIL optimizes alignment layers $\mathcal{G}(\cdot)$, while freezing the backbone networks $\mathcal{F}(\cdot)$, as illustrated in \cref{fig:trainingpipe}. By restricting training to only the alignment layer, we can afford to have large batch sizes in the contrastive loss training even for models up to 7B parameters, with 1 GPU. This is otherwise infeasible due to the combined demands of large models and batch sizes.

In our setup, paired image-text data is pre-encoded into embeddings by pretrained models only once, avoiding the need to load encoders in each forward pass. During training, only these embeddings and the lightweight alignment layer are loaded onto the GPU, significantly reducing memory requirements. This allows training on 23M examples with a single A100 GPU in ~5 hours and a batch size up to 32,768. In contrast, end-to-end contrastive training would require over 100 GPUs to handle such batch sizes \cite{clip, openclip}.

\subsection{Evaluating SAIL on Downstream Tasks.}\label{sec:evaluationsail}
In the previous section, we provided SAIL’s motivation and design choices. In this section, we consider SAIL as a foundational VLM like CLIP and evaluate the quality of the vision and language representations learned using the SAIL alignment framework. SAIL uses state-of-the-art DINOv2 as the vision model, paired with two language models: compact GTE-en-large-v1.5 (SAIL-GTE) and powerful NV-Embed-2 (SAIL-NV2). Following the optimized configurations discussed earlier, we train SAIL on a 23M Merged Dataset \cite{zheng2024dreamlip}. This dataset is a combination of CC3M, CC12M, and YFCC15M \footnote{Due to expired image URLs, we use subsets of 2.2M, 7.7M, and 12.9M images, respectively}, with high-quality captions generated from ShareGPT4.

For optimization, we use the LION optimizer \cite{lion} (with $\beta_1=0.9$, $\beta_2=0.99$), a learning rate of $10^{-5}$, and a weight decay of $10^{-7}$. We use a temperature of $t=\log 20$ and a bias of $b=-10$. The output dimension of alignment layer is 1024. Training runs for 50 epochs with a batch size of 32,768, leveraging pre-encoded embeddings to reduce memory load, using a fixed image resolution of 224.

SAIL’s performance is assessed across several zero-shot vision-language tasks, including image classification (\S \ref{sec:classification}), image-text retrieval (\S \ref{sec:retrieval}), and open-vocabulary segmentation (\S\ref{sec:segmentation}). We compare SAIL to the key baselines—CLIP and DreamCLIP \cite{zheng2024dreamlip}, both trained from scratch, as well as LiT \cite{zhai2022lit} and ShareLock \cite{sharelock}, both initialized with pretrained DINOv2 and various language models following original configurations. Additionally, we evaluate how well the visual representations learned in the SAIL framework transfer to complex vision-language tasks such as VQA (\S\ref{sec:mllm}).

\subsubsection{Zero-Shot Image Recognition}
\label{sec:classification}

We evaluated the zero-shot transfer performance of SAIL across 11 common downstream image classification tasks, and report the Top-1 accuracy for each task in \cref{tab:image classification}. With ViT-B/16 as vision architecture, we see that SAIL consistently outperforms all baselines including DreamLIP \cite{zheng2024dreamlip}, which trains from scratch, and efficient training methods like LiT \cite{zhai2022lit} and ShareLock with the same amount of paired image-text data. Notably, SAIL-B-NV2, trained on just CC12M (7.7M) image-text pairs, outperforms CLIP-B on ImageNet-1k, which was trained on the much larger LAION-400M, and achieves comparable performance in fine-grained classification tasks.

We further scale the vision model to ViT-L/14 and expand the dataset to a larger 23M merged dataset. The overall improvement from enlarging CC12M to the 23M merged dataset demonstrates the scalability of the method. SAIL-L-GTE, with a 400M-parameter language model, already achieves strong performance on ImageNet-1k, reaching 65.4\% accuracy. When equipped with a more powerful language model (SAIL-L-NV2), we observe a significant performance boost, outperforming CLIP-L on ImageNet-1k, despite using only 6\% of its training image-text paired data. SAIL also outperforms CLIP on 6 out of 10 datasets. These results underscore the pivotal role of advanced language models in enhancing vision-language tasks. In all setups, replacing GTE-en-large-v1.5 with NV-Embed-2 improves ImageNet-1k accuracy by 7-10\%. 

Interestingly, SAIL-B-NV2, with a smaller vision encoder and trained on fewer image-text pairs (7.7M), outperforms SAIL-L-GTE, which uses a larger vision encoder and 23M pairs. This demonstrates that a stronger language model (NV2 over GTE) not only boosts alignment performance but also reduces the need for extensive data.

\subsubsection{Zero-Shot Image-Text Retrieval}\label{sec:retrieval}

\begin{figure*}[!th]
    \centering
    \begin{minipage}[t]{0.65\linewidth}
        \centering
        \vfill
        \captionsetup{type=table}
        \resizebox{\linewidth}{!}{%
        \begin{tabular}{l l |cc cc| ccc|  c}
        \toprule
        & & \multicolumn{2}{c}{\cellcolor{c2!40}\textbf{MSCOCO}} & \multicolumn{2}{c|}{\cellcolor{c2!40}\textbf{Flickr30k}}  & \multicolumn{3}{c|}{\cellcolor{c3!30}\textbf{Winoground}}  & \textbf{\cellcolor{c4!30}MMVP}\\
        \textbf{Data} & \textbf{Model} & \cellcolor{c2!40}I2T & \cellcolor{c2!40}T2I & \cellcolor{c2!40}I2T & \cellcolor{c2!40}T2I & \cellcolor{c3!30}T. & \cellcolor{c3!30}I.& \cellcolor{c3!30}G. & \cellcolor{c4!30}Avg. \\
        
        \midrule
        \multicolumn{10}{c}{\textit{Model Architecture: ViT-B/16}}\\
        \multirow{6}{*}{CC12M} 
        & DreamLIP\cite{zheng2024dreamlip} & 53.3 & \cellcolor{lightorange!15}41.2 & 82.3 & \cellcolor{lightorange!15}66.6 & \cellcolor{lightorange!15}26.0 & 10.00 & 7.25 & \cellcolor{lightorange!15}24.0  \\
        & LiT\cite{sharelock} & 30.0 & 16.5 & 54.8 & 38.5 & 24.3 & 6.5 & 4.8 & - \\
        & ShareLock$\dag$\cite{sharelock} & 26.0 & 13.5 & 53.9 & 34.9 & \cellcolor{lightorange!15}26.3 & \cellcolor{lightorange!15}12.8 & 5.3 & - \\
        & SAIL-B-GTE$\dag$ & 48.2 & 37.9 & 76.5 & 63.9 & \cellcolor{lightorange!15}31.0 & \cellcolor{lightorange!15}11.5 & \cellcolor{lightorange!15}9.5 & \cellcolor{lightorange!15}23.0   \\
        & SAIL-B-NV2$\dag$ & \cellcolor{lightorange!40}57.3 & \cellcolor{lightorange!40}45.3 & \cellcolor{lightorange!40}84.1 & \cellcolor{lightorange!40}70.1 & \cellcolor{lightorange!40}35.0 & \cellcolor{lightorange!40}17.25 & \cellcolor{lightorange!40}13.0 & \cellcolor{lightorange!40}24.4 \\
        \rowcolor{gray!15}
        \textcolor{gray!70}{LAION400M} & \textcolor{gray!70}{CLIP-B} &  \textcolor{gray!70}{55.4} &  \textcolor{gray!70}{38.3} &  \textcolor{gray!70}{83.2} &  \textcolor{gray!70}{65.5} & \textcolor{gray!70}{25.7}&  \textcolor{gray!70}{11.5} &  \textcolor{gray!70}{7.75} &  \textcolor{gray!70}{19.3} \\   \midrule
        
        \multicolumn{10}{c}{\textit{Model Architecture: ViT-L/14 }}\\
        CC12M & SAIL-L-GTE & 50.4 & 39.3 & 78.4 & 66.6 & \cellcolor{lightorange!15}33.25 & \cellcolor{lightorange!15}13.0 & \cellcolor{lightorange!15}9.25 & \cellcolor{lightorange!15}17.0 \\
        23M Merged & SAIL-L-GTE & 54.1 & 42.7 & 80.8 & 68.9 & \cellcolor{lightorange!15}34.0 & \cellcolor{lightorange!15}13.25 & \cellcolor{lightorange!15}8.75 & \cellcolor{lightorange!15}22.2  \\
        
        CC12M & SAIL-L-NV2 & 57.3 & \cellcolor{lightorange!15}45.3 & 84.9 & \cellcolor{lightorange!15}73.0 & \cellcolor{lightorange!15}37.75 & \cellcolor{lightorange!15}18.25 & \cellcolor{lightorange!15}13.2 & \cellcolor{lightorange!15}28.0 \\
        23M Merged & SAIL-L-NV2 & \cellcolor{lightorange!40}62.4 & \cellcolor{lightorange!40}48.6 & \cellcolor{lightorange!40}87.6 & \cellcolor{lightorange!40}75.7 & \cellcolor{lightorange!40}40.25 & \cellcolor{lightorange!40}18.75 & \cellcolor{lightorange!40}15.0 & \cellcolor{lightorange!40}28.9 \\
      
         \rowcolor{gray!15}
        \textcolor{gray!70}{LAION400M} & \textcolor{gray!70}{CLIP-L} & \textcolor{gray!70}{59.7} & \textcolor{gray!70}{43.0} &  \textcolor{gray!70}{87.6} &  \textcolor{gray!70}{70.2} &  \textcolor{gray!70}{30.5} &  \textcolor{gray!70}{11.5} &  \textcolor{gray!70}{8.75}  &  \textcolor{gray!70}{20.0}
         \\
        
        \bottomrule
        \end{tabular}}
        \captionsetup{font=small}
\caption{\textbf{Results} on \colorbox{c2!40}{standard retrieval}, \colorbox{c3!30}{complex reasoning} and \colorbox{c4!30}{visual-centric} tasks.  We report Recall@1 for MSCOCO and Flickr30k; Text, Image and Group scores for Winoground; and the average score for MMVP. $\dag$ ViT patch size is 14.}
\label{tab: retrieval}
    \end{minipage}%
    \hfill
    \begin{minipage}[t]{0.33\linewidth}
        \centering
        \vfill
        \includegraphics[width=\linewidth]{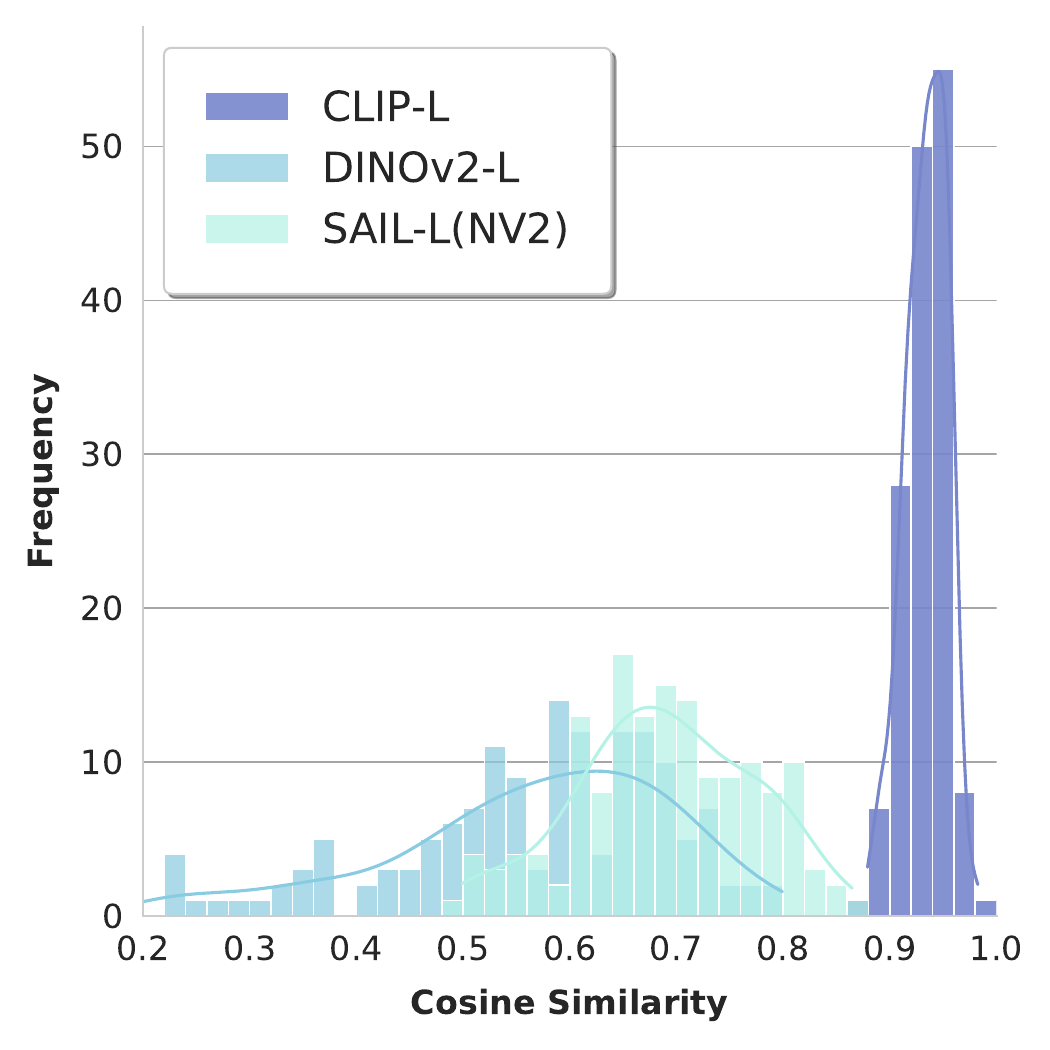} 
           \captionsetup{font=small}
        \caption{Image-Image cosine similarity distribution for 150 paired images from MMVP.}
        \label{fig:mmvp cos}
    \end{minipage}
    \vspace{-3mm}
\end{figure*}

Cross-modal retrieval is more challenging than image recognition, as it requires complex scene understanding, including spatial relationships, context, background, activities, and more. Unlike models like CLIP that lack complex reasoning capabilities \cite{thrush2022winoground, zhang2024contrasting, awal2024vismin} as they train from scratch using large, noisy image-text datasets with short captions mentioning just object names, SAIL leverages a robust pretrained language encoder, trained for complex language understanding. With minimal alignment learning on a significantly smaller image-text dataset, SAIL achieves significant cross-modal understanding improvements.

As shown in \cref{tab: retrieval}, evaluated on \colorbox{c2!40}{standard retrieval} tasks such as MSCOCO and Flickr30k, SAIL consistently outperforms other baselines, including DreamLIP, LiT, and ShareLock. It even surpasses CLIP trained on LAION-400M, while using significantly fewer samples. For example, SAIL-B-NV2 surpasses CLIP-B/16 when both are trained on the CC12M dataset. Similarly, SAIL-L-NV2, trained on 23M samples, outperforms CLIP-L/14, which is trained on 400M samples, showing particularly strong gains in text-to-image (T2I) retrieval.

SAIL also excels in \colorbox{c3!30}{complex reasoning} Winoground task. With 7.7M samples, SAIL-B-NV2 outperforms CLIP-ViT-L/14 (trained on 400M samples), underscoring the impact of NV2's advanced language understanding. As shown in  \cref{tab: retrieval}, substituting the vision model from DINOv2-B to DINOv2-L offers only marginal gains, whereas switching the language model from GTE to NV2 yields significant improvements. This highlights that complex reasoning benefits more from advanced language models than from larger vision models. Notably, SAIL-L-NV2, trained on 23M samples, achieves the best results, with an approximate 7-10\% improvement across all three metrics in Winoground compared to CLIP.  Such substantial improvements underscore SAIL’s strengths in tackling complex reasoning tasks.

On the \colorbox{c4!30}{vision-centric} MMVP benchmark \cite{tong2024eyes}, SAIL achieves strong results, outperforming CLIP trained on 400M data. Comparing SAIL(GTE) and SAIL(NV2), we observe consistent improvements. This suggests that stronger linguistic reasoning helps vision centric tasks too.


We further analyze the image-image cosine similarity on the MMVP benchmark \cite{tong2024eyes}, which consists of 150 image pairs. These pairs are selected to test subtle differences in orientation, perspective, quantity, color, and contextual details. Previous findings \cite{tong2024eyes, awal2024vismin} indicate that CLIP struggles with such distinctions, often assigning very high similarity scores even when condition varies, whereas DINOv2 effectively captures these nuances. As shown in \cref{fig:mmvp cos},  we found that SAIL’s cosine similarity distribution closely matches DINOv2’s, suggesting it retains DINOv2’s fine-grained visual acuity. By combining DINOv2’s visual precision with NV2’s linguistic depth, SAIL proves to be a powerful vision-language foundation model for nuanced visual discrimination.


\subsubsection{Open-Vocabulary Semantic Segmentation }\label{sec:segmentation}

CLIP-like models align images with sentences, allowing patch-level matching for open-vocabulary semantic segmentation \cite{maskclip, wang2023sclip}. SAIL builds on this by enhancing patch-to-label associations for segmentation by taking advantage of strong vision encoders like DINOv2. An image is represented as a sequence of tokens $X = [x_{cls}, X_{patch}]$, where $X_{patch} \in \mathbb{R}^{hw \times d}$. We compute cosine similarity between each patch and a sentence embedding $y_{text}$ (e.g., \textit{``a photo of a \{label\}''}) to produce segmentation masks following MaskCLIP \cite{maskclip}: $\mathcal{M} = \arg \max \cos(X_{patch}, y_{text})$.

We evaluated SAIL on ADE20K \cite{ade20k}, COCO-Stuff164k \cite{cocostuff}, and VOC20 \cite{voc} using mIOU to assess segmentation accuracy. As shown in \cref{tab:semantic segmentation}, SAIL outperforms baselines like CLIP, MaskCLIP \cite{maskclip}, and SCLIP \cite{wang2023sclip}, transferring DINOv2’s strong visual representations to open-vocabulary vision-language tasks and retained its fine-grained understanding capacity. This highlights SAIL’s potential for precise scene comprehension with advanced SSL models.

\begin{table}[!tb]
     \centering
     
    \resizebox{0.9\linewidth}{!}{%
      \begin{tabular}{ll ccc}
        
        \textbf{Data} & \textbf{Model (ViT-L/14)} & \textbf{ADE20K} & \textbf{Stuff} & \textbf{VOC20}  \\
        \toprule
        LAION400M &  CLIP \cite{lan2024clearclip} & 1.2 & 2.4 & 15.8 \\
        LAION400M &  MaskCLIP \cite{lan2024clearclip}& 6.9 & 8.9 & 30.1 \\
        LAION400M &  SCLIP \cite{lan2024clearclip}& 7.1 & 13.1 & 60.3 \\ \midrule
        23M Merged &  SAIL (GTE) & \cellcolor{lightorange!15}13.5 & \cellcolor{lightorange!15} 14.1 & \cellcolor{lightorange!15} 65.2\\
        23M Merged &  SAIL (NV2) & \cellcolor{lightorange!40}14.2 & \cellcolor{lightorange!40}14.7 & \cellcolor{lightorange!40}66.1\\
        
        \end{tabular}
 }

\captionsetup{font=small}
\caption{\textbf{Open-vocabulary semantic segmentation mIOU }results compared with CLIP-based methods. All models use ViT-L/14 as the vision architecture.}
\label{tab:semantic segmentation}
  \vspace{-3mm}
\end{table}

\subsubsection{Language-compatible Visual Representation}\label{sec:mllm}

\begin{figure*}[!thb]
    \centering
        \begin{minipage}[t]{0.3\linewidth}
        \centering
        \vfill
        \includegraphics[width=\linewidth]{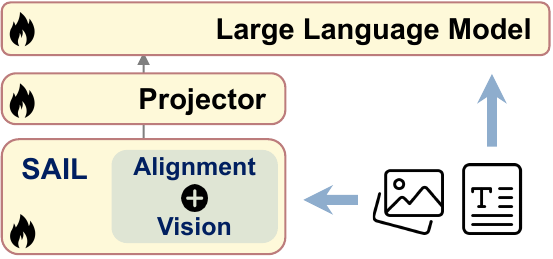} 
           \captionsetup{font=small}
        \caption{\textbf{Using SAIL's vision encoder for MLLMs.} }
        \label{fig:mllm}
    \end{minipage}
     \hfill
    \begin{minipage}[t]{0.68\linewidth}
        \centering
        \vfill
        \captionsetup{type=table}
        \resizebox{\linewidth}{!}{%
            \begin{tabular}{lcccccccc}
            \toprule
             Model@224px & VTune& SEED$^{\text{IMG}}$ & GQA & VizWiz & PoPE & TextVQA & MMB & VQA$^{\text{v2}}$ \\
            \midrule
           
            0\ \ \ DINOv2-L & \ding{55} &  61.47 & 61.08 & 44.12 &  85.5 & 45.37 & 56.96 & 74.4 \\
           
            1\ \ \ DINOv2-L & \ding{51} & 62.12 &  61.53 & 46.59 & 85.7 & 45.92 & 58.85 & 74.69\\
            2\ \ \ SAIL-L&  \ding{51} & \cellcolor{lightorange!40}65.43 & \cellcolor{lightorange!40}62.63 & \cellcolor{lightorange!40}50.00  & \cellcolor{lightorange!40}86.16 & 46.53 & 60.14 & \cellcolor{lightorange!40}76.77\\
             \rowcolor{gray!15}
             \textcolor{gray!70}{3\ \ \ CLIP-L/14$^\ast$} &  \textcolor{gray!70}{\ding{55}} &\textcolor{gray!70}{64.05} & \textcolor{gray!70}{61.58} & \textcolor{gray!70}{48.87} & \textcolor{gray!70}{85.74} & \textcolor{gray!70}{54.56} & \textcolor{gray!70}{63.06} & \textcolor{gray!70}{75.32}
 \\     
             \rowcolor{gray!15}
             \textcolor{gray!70}{4\ \ \ CLIP-L/14$^\ast$} &  \textcolor{gray!70}{\ding{51}}&\textcolor{gray!70}{64.15} & \textcolor{gray!70}{61.54} & \textcolor{gray!70}{49.93} & \textcolor{gray!70}{85.73} & \textcolor{gray!70}{54.18} & \textcolor{gray!70}{64.12} & \textcolor{gray!70}{76.36}   \\

            \bottomrule
        \end{tabular}}
        \captionsetup{font=small}
\caption{\textbf{LLaVA-1.5 with various vision models.} $^\ast$Reproduced using OpenAI CLIP-L@224 \cite{clip}. VTune indicates if the vision encoder is fine-tuned during the instruction tuning stage.}
\label{tab: mllmresults}

    \end{minipage}%

\end{figure*}

\begin{figure*}[!thb]
    \centering
    \includegraphics[width=\textwidth]{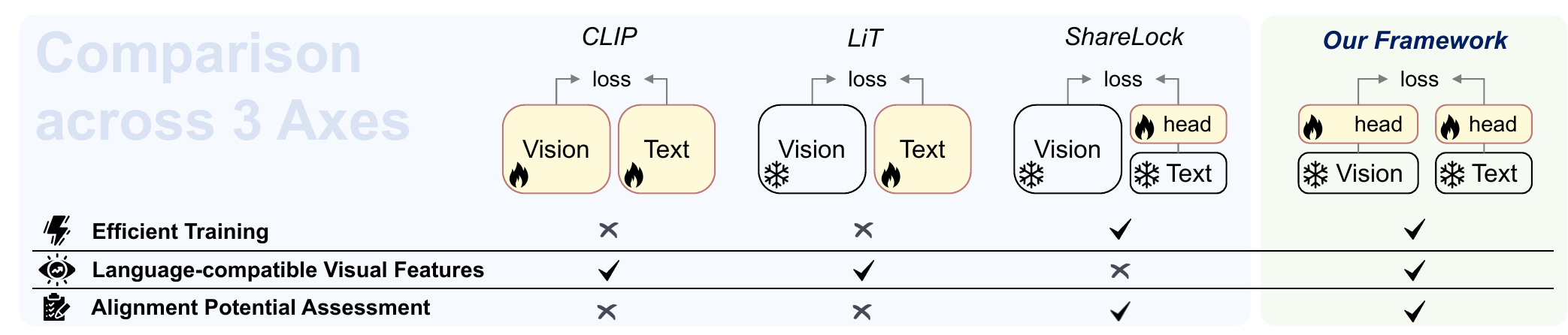}
\captionsetup{font=small}
\caption{CLIP and LiT require training full models, making them resource-intensive. LiT and ShareLock freeze vision models during training, unable to yield language-compatible visual features crucial for MLLMs. CLIP and LiT achieves modality-specific representations and cross-modal alignment simultaneously, incapable for assessing alignment potential. Our framework meets all three requirements.}
    \label{fig:comparewithotherframework}
      \vspace{-3mm}
\end{figure*}

\citet{tong2024cambrian} highlight the limitations of self-supervised vision models (e.g., DINO) as vision encoders for MLLMs, noting their lower performance across MLLM benchmarks compared to language-supervised vision models (e.g., CLIP). However, our findings demonstrate that the alignment training using SAIL framework can transform features from SSL models like DINOv2 to be more language-compatible, thus better suited for integration with MLLMs for tackling complex vision-language tasks.

We integrate SAIL-L-NV2’s vision encoder into LLaVA. SAIL's vision encoder consists of DINOv2-L and corresponding learned alignment layers. We train the model following LLaVA-1.5's training recipe \cite{liu2024improved}, and evaluate the performance of using SAIL's vision encoder in LLaVA on a range of MLLM benchmarks. \cref{tab: mllmresults} illustrates that DINOv2 benefits from fine-tuning vision encoders during the instruction-tuning stage. Thus we also fine-tune SAIL vision encoder in instruction-tuning stage as shown in \cref{fig:mllm}.

Comparing SAIL-L (row 2) with DINOv2 (row 1), SAIL (trained on 23M pairs) significantly boosts DINOv2’s performance in VQA and multimodal instruction-following tasks through alignment training, \textbf{transforming DINOv2 from lagging behind CLIP (trained on 400M pairs) to surpassing it in 5 out of 7 tasks} (rows 1-4). This comparison also includes a CLIP vision encoder that is fine-tuned during the instruction-tuning stage (row 4), highlighting that SAIL effectively learns language-compatible visual features, facilitating smoother integration with LLMs. We observe that SAIL performs poorly on TextVQA and MMB, which require Optical Character Recognition (OCR). We believe this is due to the inherent limitations of DINOv2 w.r.t OCR capabilities as we observe that the DINOv2 baselines (rows 0 and 1) also perform particularly worse than the CLIP version on these specific benchmarks.

%% file: sec/relatedwork.tex
\section{Related Work}


\paragraph{Alignment between unimodal models.}
Recent studies reveal that alignment can emerge within unimodal models even without explicitly aligning them with each other. \citet{huh2024platonic} use mutual nearest-neighbor metrics to suggest that models across modalities align to a shared statistical reality. \citet{maniparambil2024vision} find that vision encoders  exhibit high semantic similarity with language encoders using centered kernel distance. While these studies imply inherent alignment within unimodal models, they rely on proxy measurements without directly assessing cross-modal distance for individual image-text pairs. 
\citet{moschella2022relative} introduce relative representation, encoding samples based on similarity to a fixed set of anchors, ASIF \cite{norelli2023asif} further extends this for cross-modal retrieval, requiring no training.
 Our approach directly measures the cross-modal distance for individual image-text pairs to quantitatively examine how modality-specific features impact alignment. 


\paragraph{Efficient tuning} of foundational VLMs like CLIP has remained underexplored. One line of work improves data efficiency by using improved captions from LLMs \cite{lalip} or MLLMs \cite{zheng2024dreamlip}, though computational demands remain high. Other methods combine pre-trained unimodal models to reduce data and compute needs. For instance, LiT \cite{zhai2022lit} aligns a frozen vision model with a language model trained from scratch, which still requires substantial effort. \citet{khan2023contrastive} explore parameter-efficient contrastive alignment by tuning layer norms and bias terms; ShareLock \cite{sharelock} further introduces a tunable head over the frozen language model to achieve alignment with fewer parameters. However, these methods have limited alignment performance and do not improve the base vision encoder, restricting the ability to transfer improved vision-language alignment capabilities to MLLMs. As shown in \cref{fig:comparewithotherframework}, our framework stands out by meeting three key points compared to other frameworks: efficient training, learning language-compatible visual features, and a more direct method for assessing alignment potential, consistent with how inference is done with such VLMs.


%% file: sec/conclusion.tex
\section{Conclusion}

We propose a probing framework to evaluate cross-modal alignment between pretrained unimodal vision and language models and to explore how modality-specific features impact this alignment. We show that the clustering quality of SSL features is crucial for alignment and that advanced language models are essential for reasoning in vision-language tasks. We present SAIL, a framework that efficiently learns alignment with less data and compute demands. SAIL excels in zero-shot classification, retrieval, complex reasoning, and segmentation, outperforming CLIP trained on 400M image-text pairs with only 23M data. This study highlights the potential of efficient alignment strategies for advancing vision-language integration, aiming to accelerate research for resource-limited academic groups on topics such as architecture, losses, and data.

%% file: sec/X_suppl.tex
\clearpage
\setcounter{page}{1}
\maketitlesupplementary

\section{Reproducibility Statement}
To ensure the reproducibility of our work, we are committed to making all training code, datasets, and model weights publicly available at \href{https://lezhang7.github.io/sail.github.io/}{Project Page}. Detailed documentation will accompany the codebase to facilitate easy replication of our experiments. Hyperparameter settings, training configurations, and any preprocessing steps will also be thoroughly outlined. By providing these resources, we aim to promote transparency, enable future research, and support the broader community in building upon our work.

\section{Alignment Assessment Training Details} \label{sec:alignmentinvest}

\paragraph{Alignment Probing}
The alignment probing method uses contrastive learning to train linear layers, referred to as alignment layers, for aligning pretrained unimodal vision and language representation spaces. 

Specifically, with a \textbf{frozen} image encoder \(\mathcal{F}_I(\cdot)\) and a \textbf{frozen} text encoder \(\mathcal{F}_T(\cdot)\), the corresponding \textbf{linear} layers \(\mathcal{G}_I(\cdot)\) and \(\mathcal{G}_T(\cdot)\) are trained using the refined sigmoid loss on CC3M dataset with ShareGPT4-enhanced captions and incorporating the multiple positive caption contrast, as described in \cref{sec:cheaptrain}. For optimization, we use the LION optimizer (with $\beta_1=0.9$, $\beta_2=0.99$), a learning rate of $10^{-5}$, and a weight decay of $10^{-7}$. We use a temperature of $t=\log 20$ and a bias of $b=-10$. The output dimensionality of the linear layer (alignment dimensionality) is 2048. Training runs for 100 epochs with a batch size of 32,768, using a fixed image resolution of 224.





\begin{figure}[!htb]
    \centering
    \includegraphics[width=1\linewidth]{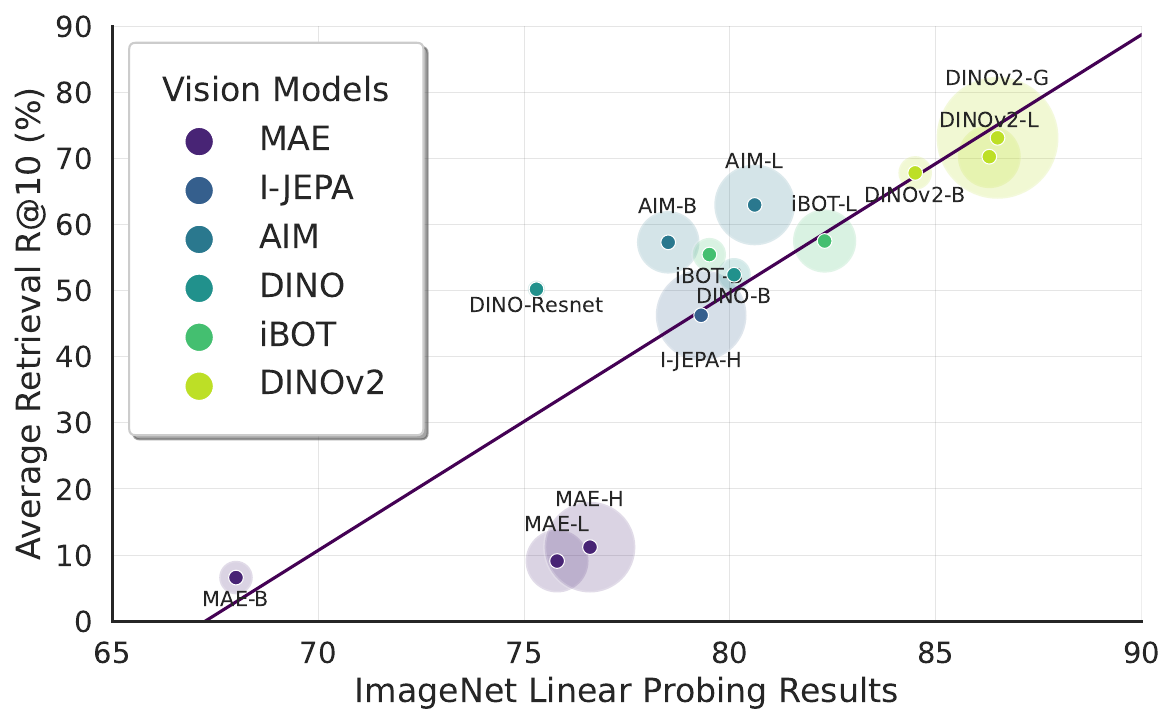}
    \caption{\textbf{Linear alignment probing results} between Imagenet linear probing accuracy and average retrieval R@10 (our metric). MAE serves as an outlier, achieving high linear probe performance but low alignment performance. }
    \label{fig:visionalign_linearprobe}
\end{figure}

\paragraph{Linear Probing vs. Alignment scores} \cref{fig:visionalign_linearprobe} illustrates the relationship between our alignment metric and the ImageNet linear probing classification accuracy of models. Compared to kNN (refer to \cref{fig:visionalign1}), the linear correlation between these two metrics is weaker, with a Pearson correlation coefficient of 0.847. This highlights that non-linear separability (i.e., clustering quality) matters more than linear-separability for image-text alignment.

  \begin{table*}[!thb]
        \centering
        \resizebox{\linewidth}{!}{%
        \begin{tabular}{l l |cc cc| ccc|  c| cc}
        \toprule
        & & \multicolumn{2}{c}{\cellcolor{c1!40}\textbf{MSCOCO}} & \multicolumn{2}{c|}{\cellcolor{c1!40}\textbf{Flickr30k}}  & \multicolumn{3}{c|}{\cellcolor{c2!40}\textbf{Winoground}}  & \textbf{\cellcolor{c3!30}MMVP} & \textbf{\cellcolor{c4!30}ImageNet} & \textbf{\cellcolor{c4!30}10 Classification}\\
        \textbf{Data} & \textbf{Model} & \cellcolor{c1!40}I2T & \cellcolor{c1!40}T2I & \cellcolor{c1!40}I2T & \cellcolor{c1!40}T2I & \cellcolor{c2!40}T. & \cellcolor{c2!40}I.& \cellcolor{c2!40}G. & \cellcolor{c3!30}10 Avg. &  \cellcolor{c4!30}Top1. &  \cellcolor{c4!30}Avg. \\
        
        \midrule
        \multicolumn{12}{c}{\textit{Model Architecture: ViT-B/16}}\\
        \multirow{6}{*}{CC12M} 
        & DreamLIP & 53.3 & \cellcolor{lightorange!15}41.2 & 82.3 & \cellcolor{lightorange!15}66.6 & \cellcolor{lightorange!15}26.0 & 10.00 & 7.25 & \cellcolor{lightorange!15}24.0 & 50.3 & 49.9  \\
        & LiT$\ddag$ & 30.0 & 16.5 & 54.8 & 38.5 & 24.3 & 6.5 & 4.8 & - & 56.2 & - \\
        & ShareLock(Llama3)$\dag$$\ddag$ & 26.0 & 13.5 & 53.9 & 34.9 & \cellcolor{lightorange!15}26.3 & \cellcolor{lightorange!15}12.8 & 5.3 & - & 59.1 & - \\
         & ShareLock(NV2)$\dag$ & 39.6 & 23.1 & 68.1 & 49.3 &  \cellcolor{lightorange!15}33.25 & \cellcolor{lightorange!15}13 & \cellcolor{lightorange!15}9.75 & 15.56 & 61.9 & 62.0  \\
        & SAIL-B (GTE)$\dag$ & 48.2 & 37.9 & 76.5 & 63.9 & \cellcolor{lightorange!15}31.0 & \cellcolor{lightorange!15}11.5 & \cellcolor{lightorange!15}9.5 & \cellcolor{lightorange!15}23.0 & 58.7 & 57.7   \\
        & SAIL-B (NV2)$\dag$ & \cellcolor{lightorange!40}57.3 & \cellcolor{lightorange!40}45.3 & \cellcolor{lightorange!40}84.1 & \cellcolor{lightorange!40}70.1 & \cellcolor{lightorange!40}35.0 & \cellcolor{lightorange!40}17.25 & \cellcolor{lightorange!40}13.0 & \cellcolor{lightorange!40}24.4 & \cellcolor{lightorange!40}68.1 & 65.4 \\
        \rowcolor{gray!15}
        \textcolor{gray!70}{LAION400M} & \textcolor{gray!70}{CLIP-B} &  \textcolor{gray!70}{55.4} &  \textcolor{gray!70}{38.3} &  \textcolor{gray!70}{83.2} &  \textcolor{gray!70}{65.5} & \textcolor{gray!70}{25.7}&  \textcolor{gray!70}{11.5} &  \textcolor{gray!70}{7.75} &  \textcolor{gray!70}{19.3} & \textcolor{gray!70}{67} & \textcolor{gray!70}{65.5} \\  

        \midrule
        \multicolumn{12}{c}{\textit{Model Architecture: ViT-L}}\\
         23M Merged& SAIL-L (NV2)$\dag$ & \cellcolor{lightorange!40}62.4 & \cellcolor{lightorange!40}48.6 & \cellcolor{lightorange!40}87.6 & \cellcolor{lightorange!40}75.7 & \cellcolor{lightorange!40}40.25 & \cellcolor{lightorange!40}18.75 & \cellcolor{lightorange!40}15.0 &
         \cellcolor{lightorange!40}28.9 & \cellcolor{lightorange!40}73.4 & 72.1 \\

           \rowcolor{gray!15}
        \textcolor{gray!70}{LAION400M} & \textcolor{gray!70}{CLIP-L} &  \textcolor{gray!70}{59.7} &  \textcolor{gray!70}{43.0} &  \textcolor{gray!70}{87.6} &  \textcolor{gray!70}{70.2} & \textcolor{gray!70}{30.5}&  \textcolor{gray!70}{11.5} &  \textcolor{gray!70}{8.75} &  \textcolor{gray!70}{20.0}& \textcolor{gray!70}{72.7} & \textcolor{gray!70}{75.9}  \\

        \bottomrule
        \end{tabular}}
        \captionsetup{font=small}
\caption{\textbf{Results} on \colorbox{c1!40}{standard retrieval}, \colorbox{c2!40}{complex reasoning}, \colorbox{c3!30}{visual-centric}, and \colorbox{c4!40}{classification} tasks. We report Recall@1 for MSCOCO and Flickr30k, Text, Image, and Group scores for Winoground, and the average score across 9 visual patterns for MMVP. $\ddag$ indicates cited results, and $\dag$ denotes a ViT patch size of 14. 10 Classification tasks include: Food101, CIFAR10, CIFAR100, SUN397, Cars, Aircraft, DTD, Pets, Caltech101, and Flowers.}
\label{tab: additional results}
    \end{table*}%

\section{Additional Comparison with ShareLock}

In \cref{sec:classification} and \cref{sec:retrieval}, we compare our method directly with the concurrent work ShareLock, using the reported results from \cite{sharma2024vision}, as the code was not open-source at the time of submission. ShareLock utilizes the LLaMA3-8B as the language encoder, which differs from the NV-Embed-2 language encoder used in SAIL. To ensure a fair comparison, we reproduced ShareLock's results after consulting the authors, using the same vision and language backbones as SAIL (DINOv2-B and NV-Embed-2). We adhered strictly to the training details provided in the original paper \cite{sharma2024vision} and present evaluation results for classification and retrieval tasks in \cref{tab: additional results}.

\begin{figure}
    \centering
    \includegraphics[width=0.7\linewidth]{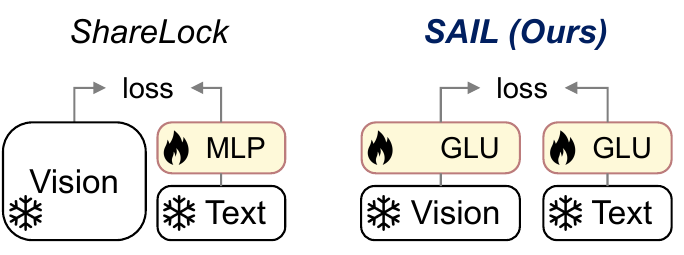}
    \caption{\textbf{Method comparison.} SAIL shows consistent improved performance over ShareLock.}
    \label{fig:sharelocksail}
\end{figure}

The ShareLock results demonstrates that replacing LLaMA3-8B with NV-Embed-2 significantly improves alignment performance across benchmarks. Also we see that, using the same vision and language backbones, SAIL (NV2) consistently outperforms ShareLock (NV2) across all tasks by a significant margin. This highlights the effectiveness of incorporating alignment layers for both vision and language models (see \cref{fig:sharelocksail} for differences), as well as the advantages of our proposed optimized training methodologies.

\paragraph{Experiments without enhanced caption}

We provide additional results in \cref{tab:wocaption} by training SAIL on CC12M raw dataset without long captions.  We see that SAIL still significantly outperforms both ShareLock and LiT across all benchmarks, further validating the effectiveness of our method.

\begin{table*}[!thb]
    \centering
    \resizebox{\linewidth}{!}{%
    \begin{tabular}{l l |cc cc| ccc| c| cc}
    \toprule
    & & \multicolumn{2}{c}{\cellcolor{c1!40}\textbf{MSCOCO}} & \multicolumn{2}{c|}{\cellcolor{c1!40}\textbf{Flickr30k}}  & \multicolumn{3}{c|}{\cellcolor{c2!40}\textbf{Winoground}}  & \textbf{\cellcolor{c3!30}MMVP} & \textbf{\cellcolor{c4!30}ImageNet} & \textbf{\cellcolor{c4!30}10 Classification} \\
    \textbf{Data} & \textbf{Model (DINOv2-B)} & \cellcolor{c1!40}I2T & \cellcolor{c1!40}T2I & \cellcolor{c1!40}I2T & \cellcolor{c1!40}T2I & \cellcolor{c2!40}T. & \cellcolor{c2!40}I.& \cellcolor{c2!40}G. & \cellcolor{c3!30}10 Avg. &  \cellcolor{c4!30}Top1. &  \cellcolor{c4!30}Avg. \\
    \midrule

    CC12M raw& LiT$\ddag$ & 30.0 & 16.5 & 54.8 & 38.5 & 24.3 & 6.5 & 4.8 & - & 56.2 & - \\
    CC12M raw & ShareLock (NV2) & 39.6 & 23.1 & 68.1 & 49.3 &  \cellcolor{lightorange!15}33.25 & \cellcolor{lightorange!15}13 & \cellcolor{lightorange!15}9.75 & 15.56 & 61.9 & 62.0  \\
    CC12M raw& SAIL-B (NV2) & \cellcolor{lightorange!40}45.6 & \cellcolor{lightorange!40}32.9 & \cellcolor{lightorange!40}74.2 & \cellcolor{lightorange!40}60.6 & \cellcolor{lightorange!40}35.0 & \cellcolor{lightorange!40}19.0 & \cellcolor{lightorange!40}14.25 & \cellcolor{lightorange!40}25.2 & \cellcolor{lightorange!40}69.2 & \cellcolor{lightorange!40}66.4 \\

    \bottomrule
    \end{tabular}}
    \captionsetup{font=small}
    \caption{Trained on CC12M raw captions.}
    \label{tab:wocaption}
\end{table*}

\section{Dataset used for evaluation}

\paragraph{Winoground evaluation}

\begin{figure}[!htb]%
    \centering
    \subfloat[\centering some plants surrounding a lightbulb]{{\includegraphics[height=3cm]{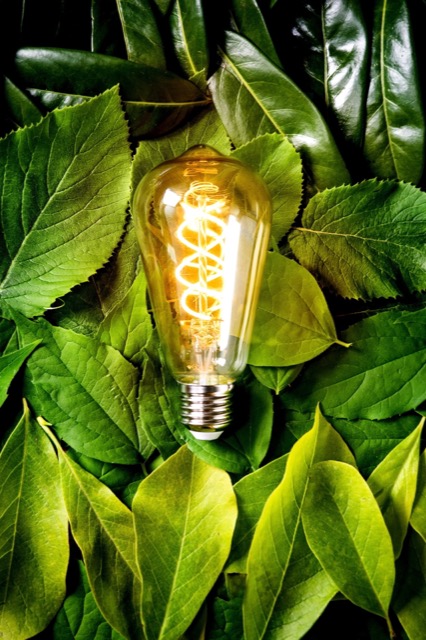} }}%
    \qquad
    \subfloat[\centering a lightbulb surrounding some plants]{{\includegraphics[height=3cm]{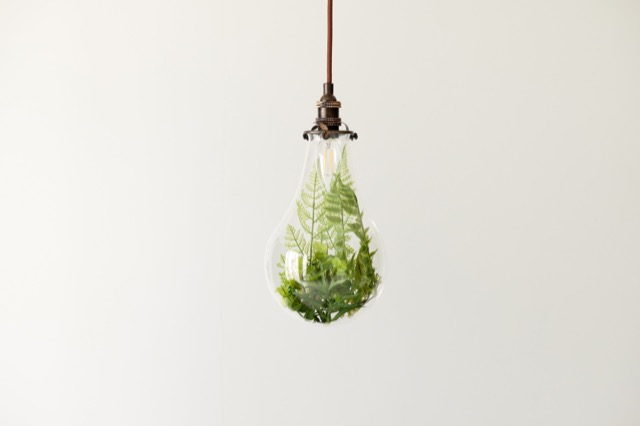} }}%
    \caption{An example from Winoground. }%
    \label{fig:winogroundexample}%
\end{figure}

Winoground \cite{thrush2022winoground} is a benchmark designed to evaluate the ability of vision-language models to perform visio-linguistic compositional reasoning, we provide one example as in \cref{fig:winogroundexample}. The task involves matching the correct image-captions pairs given two images and two captions, where the captions contain identical sets of words but in different orders, requiring fine-grained reasoning about the visual and textual alignment.

Performance is measured using three metrics: text score, image score, and group score, defined as follows. Given two image-text pairs \((I_0, T_0)\) and \((I_1, T_1)\), and a similarity function \(s(\cdot)\) provided by the model:

The \textbf{text score }evaluates if the ground-truth caption for each image is scored higher than the alternative caption. It is computed as:
\begin{equation}
f\left(T_0, I_0, T_1, I_1\right)= \begin{cases}1 \quad & \text { if } s\left(T_0, I_0\right)>s\left(T_1, I_0\right) \\ & \text { and } s\left(T_1, I_1\right)>s\left(T_0, I_1\right) \\ 0 & \text { otherwise }\end{cases}
\end{equation}

The \textbf{image score} tests whether the correct image is selected for each caption. It is computed as:
\begin{equation}
g\left(T_0, I_0, T_1, I_1\right)= \begin{cases}1 \quad & \text { if } s\left(T_0, I_0\right)>s\left(T_0, I_1\right) \\ & \text { and } s\left(T_1, I_1\right)>s\left(T_1, I_0\right) \\ 0 & \text { otherwise }\end{cases}
\end{equation}

The \textbf{group score} combines the two previous metrics, requiring both to be correct simultaneously:
\begin{equation}
h\left(T_0, I_0, T_1, I_1\right)= \begin{cases}1 \quad & \text { if } f\left(T_0, I_0, T_1, I_1\right) \\ & \text { and } g\left(T_0, I_0, T_1, I_1\right) \\ 0 & \text { otherwise }\end{cases}
\end{equation}

These metrics collectively assess whether the model can align text and images accurately while reasoning over compositional semantics.

\paragraph{MLLM benchmarks} In \cref{sec:mllm}, we combine SAIL vision encoder with LLaVA-1.5 and evaluated on various downstream VQA and instruction-following benchmarks. Below we provide a description of each of these benchmarks.

\begin{itemize}[leftmargin=2em]
    \item \textbf{SEED} \cite{li2023seed}: SEED-Bench offers a comprehensive evaluation framework with \textbf{19K multiple-choice questions}, featuring accurate human annotations—six times larger than existing benchmarks. It spans \textbf{12 evaluation dimensions}, covering comprehension in both \textbf{image} and \textbf{video modalities}. The use of multiple-choice questions with human-annotated ground truth answers ensures \textbf{objective and efficient model assessment}, removing the need for human or GPT intervention during evaluation.

    \item \textbf{GQA} \cite{hudson2019gqa}: GQA stands out as a \textbf{dataset for real-world visual reasoning} and compositional question answering, addressing key limitations of earlier VQA datasets. It emphasizes \textbf{reasoning, compositionality}, and the \textbf{grammar-based generation} of natural language queries, pushing models to engage in structured and logical visual understanding.

    \item \textbf{VizWiz} \cite{gurari2018vizwiz}: VizWiz features over \textbf{31,000 visual questions} originating from visually impaired individuals who used mobile phones to capture images and record spoken queries. Each question is paired with \textbf{10 crowdsourced answers}, introducing challenges such as \textbf{blurry images, partial scenes}, and diverse visual content, providing a real-world perspective on VQA.

    \item \textbf{PoPE} \cite{pope}: PoPE targets \textbf{Object Hallucination} in multimodal large language models (MLLMs) by focusing on challenging visual reasoning tasks. It transforms hallucination evaluation into a \textbf{binary classification task}, using \textbf{Yes-or-No questions} about specific objects (e.g., ``Is there a car in the image?''), offering a direct and interpretable measure of model accuracy in visual interpretation.

    \item \textbf{TextVQA} \cite{textvqa}: TextVQA challenges models to \textbf{extract and reason about textual information} embedded in images, such as names, prices, and other details. It heavily relies on \textbf{Optical Character Recognition (OCR)} to parse diverse and complex text inputs. The dataset pushes OCR systems to handle variations in \textbf{font styles, sizes, orientations}, and noisy scenes, providing critical inputs for downstream reasoning tasks.

    \item \textbf{MMBench} \cite{liu2025mmbench}: MMBench is a \textbf{systematically designed benchmark} for evaluating the diverse abilities of large vision-language models (VLMs). It includes \textbf{3,000+ multiple-choice questions} across \textbf{20 ability dimensions}, such as \textbf{object localization} and \textbf{social reasoning}. Each dimension is represented by \textbf{125+ balanced questions}, ensuring robust evaluation. Tasks such as text interpretation within images further emphasize the importance of \textbf{OCR capabilities} in vision-language modeling.

    \item \textbf{VQAv2} \cite{vqav2}: As one of the most widely used benchmarks for VQA, VQAv2 introduces \textbf{balanced questions} to mitigate language biases. It emphasizes \textbf{visual reasoning}, requiring models to align language understanding with accurate visual grounding, setting a strong standard for comprehensive VQA tasks.
\end{itemize}

\section{Pre-encoding Efficiency} \label{sec:preencode}
The SAIL training pipeline comprises two key stages: pre-encoding and alignment tuning. Here, we provide an estimate of the pre-encoding speed for models used in constructing SAIL. Encoding speed is influenced by factors such as hardware capabilities, model architecture, and the availability of acceleration techniques like FlashAttention. Additionally, for language models, sentence length significantly affects encoding performance.

Since encoding times depend on hardware and model configurations, we report approximate times based on our training setup, utilizing a single A100-80G GPU:

\begin{itemize}
    \item With DINOv2-L using scaled dot-product attention, encoding 224x224 resolution images from CC3M achieves a throughput of approximately \(\sim 830\) samples/s.
    \item For GTE-en-large-v1.5 with FlashAttention, the throughput is \(\sim 2350\) samples/s for short raw captions and \(\sim 130\) samples/s for longer, high-quality captions (truncated to a maximum of 1024 tokens).
    \item With NV-embed-2, the throughput is \(\sim 170\) samples/s for short raw captions and \(\sim 25\) samples/s for longer, high-quality captions (truncated to a maximum of 1024 tokens).
\end{itemize}

With acceleration methods such as FlashAttention and vLLM, the encoding speed could be further enhanced for these models. Note that encoding is performed only once and reused multiple times during training.